\documentclass[conference]{IEEEtran}
\IEEEoverridecommandlockouts

\usepackage{cite}
\usepackage{amsmath,amssymb,amsfonts}
\usepackage{algorithmic}
\usepackage{graphicx}
\usepackage{textcomp}
\usepackage{url}
\usepackage{multirow}
\usepackage[table,xcdraw]{xcolor}
\usepackage{hyperref} 
\usepackage{cleveref}
\def\BibTeX{{\rm B\kern-.05em{\sc i\kern-.025em b}\kern-.08em
    T\kern-.1667em\lower.7ex\hbox{E}\kern-.125emX}}
\usepackage{comment}
\begin{document}

\bstctlcite{IEEEexample:BSTcontrol}	

\renewcommand{\IEEElabelindentfactori}{0}

\title{Adversarial Robustness of Bottleneck Injected Deep Neural Networks for Task-Oriented Communication
}

\author{
    \IEEEauthorblockN{Alireza Furutanpey, 
    Pantelis A. Frangoudis,
    Patrik Szabo,
    Schahram Dustdar}
        TU Wien \\
        Distributed Systems Group       
}

\maketitle

\begin{abstract}
This paper investigates the adversarial robustness of Deep Neural Networks (DNNs) using Information Bottleneck (IB) objectives for task-oriented communication systems. We empirically demonstrate that while IB-based approaches provide baseline resilience against attacks targeting downstream tasks, the reliance on generative models for task-oriented communication introduces new vulnerabilities. Through extensive experiments on several datasets, we analyze how bottleneck depth and task complexity influence adversarial robustness. Our key findings show that Shallow Variational Bottleneck Injection (SVBI) provides less adversarial robustness compared to Deep Variational Information Bottleneck (DVIB) approaches, with the gap widening for more complex tasks. Additionally, we reveal that IB-based objectives exhibit stronger robustness against attacks focusing on salient pixels with high intensity compared to those perturbing many pixels with lower intensity. Lastly, we demonstrate that task-oriented communication systems that rely on generative models to extract and recover salient information have an increased attack surface. The results highlight important security considerations for next-generation communication systems that leverage neural networks for goal-oriented compression.
\end{abstract}

\section{Introduction} \label{sec:introduction}
Intelligent tasks refer to programs that classical control structures cannot compute tractably or with sufficient precisions, which is common in visual applications, such as image recognition. Deep Learning (DL) has repeatedly demonstrated that it can solve recognition tasks reliably. Unsurprisingly, applications with stringent performance criteria (e.g., remote sensing, video analytics) increasingly offload requests to a remotely deployed large Deep Neural Network (DNN).
The pervasiveness of DNNs exposes significant vulnerabilities to adversarial attacks~\cite{amlsurvey}. Another limitation is that continuous offloading of high-dimensional visual data must compete for limited bandwidth, which may lead to network congestion. Task-oriented communication~\cite{Gunduz23} has emerged as a paradigm to meet the need for solving intelligent tasks. Compression for task-oriented communication uses a semantic rate-distortion objective to transmit only the most salient bits. Among the earliest examples is the information bottleneck~\cite{informationbottleneck} (IB), which is still the foundation of modern approaches. Notably, IB-based objectives improve adversarial robustness for DNN predictors~\cite{deepinformationbottleneck}. The idea is that perturbations are intrinsically redundant information, and the IB objective naturally enhances the robustness of DNNs by learning to discard redundancy more aggressively along their information processing path~\cite{shwartz2017opening}. However, the established consensus on the value of IB-based objectives is based on the assumption that the networks are \textit{deep}. Yet, task-oriented communication is feasible only when paired with lightweight compression such that the codec computational overhead is offset by the reduced transmission costs~\cite{Mostaani24, Peng24}. Moreover, there are additional constraints on the encoder design. While a neural encoder may still be wide enough to leverage parallelization from onboard AI accelerators, meeting stringent latency requirements demands reducing the number of sequential operations. Hence, envisioned future communication networks that rely on neural encoding schemes will realistically converge towards \textit{shallow} networks. 

To this end, this work investigates the robustness of methods that apply an IB-based objective intended for task-oriented communication. 
Specifically, we apply several common adversarial attacks on recent approaches based on \textit{Shallow Variational Bottleneck Injection} (SVBI)~\cite{frankensplit, fool, matsubara2022supervised, garbo, condar}).  SVBI focuses on information necessary only for practically relevant tasks by targeting the shallow representation of foundational models as a reconstruction target in the rate-distortion objective.  Our results show that deep networks trained with a traditional IB objective exhibit higher adversarial robustness than SVBI. However, a shallow variational encoder still provides a defense mechanism that results in considerably more robust models than purely discriminative models trained with non-IB objectives. We finalize our experiments by accentuating the increased attack surface of systems that rely on generative models for communicating salient information with a simple attack specifically targeting generative models. In other words, the overall system is more vulnerable even if task-oriented communication is intrinsically more robust than passing messages through conventional channels for downstream tasks. 
We summarize our contributions as follows:
\begin{itemize}
    \item Empirically demonstrating that task-oriented communication systems have an increased attack surface.
    \item Showing that adversarial robustness for task-oriented communication systems requires a study distinct from general research on security for DNNs.
    \item Determining the role of bottleneck depth for adversarial robustness with IB-based objectives.
\end{itemize}
We hope our results and insights can facilitate research in securing next-generation communication systems that rely on otherwise easily exploitable neural networks. 

\section{Background \& Related Work} \label{sec:background}
\subsection{Adversarial Attacks on DNNs} \label{subsec:attackdesc}
Adversarial attacks represent a significant challenge for deploying AI systems. The susceptibility of DNNs to adversarial examples was first investigated by Szegedy et al. \cite{szegedy2014intriguing}, who demonstrated that small, imperceptible perturbations to input data can lead to significant misclassifications. Adversarial attacks are classified into white-box and black-box attacks. White-box attacks assume complete model knowledge, including architecture and gradient calculation, allowing for computing highly effective adversarial samples.  In contrast, black-box attacks assume no access to model details and are generally more challenging but more realistic for real-world scenarios. The following briefly describes the attacks we have chosen due to their influence and being subject to numerous follow-up studies.  The focus is on white-box attacks due to the open nature of ML research and the popularity of readily available open-source weights for a wide range of tasks. 

\subsubsection{Fast Gradient Sign Method (FGSM)} \label{subsubsec:fgsm}
 FGSM by Goodfellow et al.~\cite{goodfellow2015explaining} efficiently generates adversarial examples by leveraging the gradient of the loss function. FGSM adjusts the input along the gradient’s direction, with the perturbation defined as:
\begin{equation}
x_{a d v}=x+\epsilon \cdot \operatorname{sign}\left(\nabla_x J(x, y)\right)
\end{equation}

\noindent where \( x \) is the input, \( \epsilon \) controls perturbation magnitude, and \( \nabla_x J(x, y) \) is the gradient of the loss concerning the input. 
\subsubsection{Carlini and Wagner (C \& W)}
The attack by Carlini and Wagner~\cite{carlini2017towards} minimizes the  $L_2$, $L_0$, or $L_{\infty}$ distance between the original input and the adversarial example, and a term that penalizes classifications other than the desired target class using the objective function:
\begin{equation}
    J(x^{\prime}) = \alpha \cdot || x - x^{\prime} ||_{p} + \beta \cdot \mathcal{L}_{\text{mcls}}(f(x^{\prime}), y_{t})
\end{equation}

\noindent where $x^{\prime}$ is the perturbed input, $\alpha, \beta$ balance the terms, and $\mathcal{L}_{\text{mcls}}$ is the missclassification loss.  Notably, this attack is shown to be highly effective against networks pre-trained on ImageNet, which are commonly used to finetune by practical recognition tasks. 
\subsubsection{Elastic-Net Attacks on DNNs (EAD)} \label{subsubsec:ead}
This method~\cite{Chen2017EADEA} is particularly useful for producing sparse perturbations, which can trick DNNs while maintaining minimal changes to the input. It generates adversarial samples by minimizing the objective
\begin{equation}
    c \cdot f(\mathbf{x}, t)+\beta\left\|\mathbf{x}-\mathbf{x}_0\right\|_1+\left\|\mathbf{x}-\mathbf{x}_0\right\|_2^2
\end{equation}

\noindent where $f (x, t)$ is a target loss function and $c, \beta \geq 0$ are the regularization parameters.
EAD's dual-norm optimization is an interesting alternative benchmark for evaluating how variational bottleneck injection handles diverse attack strategies.
\subsubsection{Jacobian-based Saliency Map Attack (JSMA)} \label{subsubsec:jsma}
The JSMA attack by Papernot et al.~\cite{Papernot2015TheLO} constructs adversarial examples by identifying and perturbing input features most critical to the classifier's decision-making process. Unlike gradient-based methods, JSMA uses forward derivatives to create a saliency map, guiding perturbations to specific input features. Given that variational bottleneck techniques may alter feature representations, testing JSMA will allow us to explore how bottleneck injection influences feature saliency and adversarial resilience.
\subsubsection{Targeting Generative Models} \label{subsubsec:tabacof}
Lastly, we include the attack introduced by Tabacof et al. \cite{Tabacof2016AdversarialIF} to demonstrate the increased attack surface of communication systems that deploy generative models. This attack aims to disrupt reconstruction and induce the encoder to produce a completely different target image. This would undermine the potential defensive role of autoencoders in de-noising classifier inputs. Note that the efficacy of the attack towards the autoencoder is irrespective of whether we map the latent to an approximation of the original image (i.e., reconstruction) or use it for some image recognition downstream task~\cite{fool}. 

\subsection{Information Bottleneck in Task-Oriented Compression}
Using Shannon’s rate-distortion (r-d) theory~\cite{shannon1959coding}, we seek a mapping bound by a distortion constraint from a random variable (r.v.) $X$ to a r.v. $Y$, minimizing the bitrate of the outcomes of $X$. More formally, given a distortion measure $\mathcal{D}$ and a distortion constraint $D_c$, the minimal bitrate is characterized by the \textit{rate-distortion function}:
\begin{equation}
\underset{P_{Y|X}}{\mathrm{min}}\; I(X;Y)\; \text{s.t.}\; \mathcal{D}(X,Y) \leq D_c\label{eq:rdbasics}
\end{equation}
where $I(X;Y)$ is the mutual information and is defined as
\begin{equation}
\mathrm{I}(X;Y)=\int_{}\int_{} p(x,y) \log \left(\frac{p(x,y)}{p(x) p(y)}\right) dxdy\label{eq:mutualinf}
\end{equation}
In lossy image compression, $Y$ is typically an approximate reconstruction of $X$. This objective lends itself to the Information Bottleneck that maps $X$ to a hidden representation $Z$, which is minimally informative of $X$ but is also maximally informative about a target prediction task $Y$. In other words, it is essentially a flavor of the lossy-source coding problem using a different loss as a fidelity measure for the distortion constraint.

\subsubsection{Deep Variational Information Bottleneck (DVIB)}
Given ground-truth labels $Y$ from a joint distribution $P_{X,Y}$, the Deep Variational Information Bottleneck objective is to maximize
\begin{equation}
I(Z;Y) - \beta I(Z;X)\label{eq:vanillavib}
\end{equation}
where $\beta$ is a Lagrange multiplier. 
To approximate $I(Z;Y)$ we can apply the conditional cross entropy (CE) $H(P_Y, P_{\tilde{Y}|Z})$.
The first term is commonly referred to as the \textit{relevance} and the second as the \textit{complexity}. While the original work~\cite{deepinformationbottleneck} considers the complexity term a regularizer, Singh et al.~\cite{singh2020} apply it as a rate term to end-to-end train a neural compression model. Dubois et al.~\cite{dubois2021} generalize the information bottleneck objective for compression that preserves salient pixels for a set of tasks that share common properties. 
However, both works rely on deep networks and place the bottleneck at the penultimate layer or by first passing the input to a large pre-trained encoder.
\subsubsection{Shallow Variational Bottleneck Injection (SVBI)} \label{subsubsec:svbi}
\begin{figure}[htb]
    \centering
    \includegraphics[width=\columnwidth]{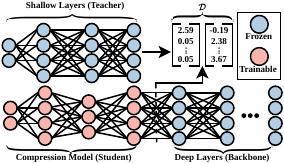}
    \caption{Head Distillation Loss: The shallow features of pre-trained large models are cut and used as a teacher network to train the compression model.}
    \label{fig:hd}
\end{figure}

Instead of targeting a particular task $Y$, SVBI 
considers a foundational model $\mathcal{M}$, that supports a set of unknown tasks $\mathcal{Y} = \{Y_1, Y_2, \dots, Y_t\}$. Moreover, it partitions $\mathcal{M}$ into two disjoint sets of shallow and deep layers $\mathcal{M} = {(\mathcal{T}, \mathcal{H}})$, such that for observations $X$, $\mathcal{M}(X) = \mathcal{T}(\mathcal{H}(X))$, and $\mathcal{H} = H$ is a \textit{shallow} hidden representation of $X$. Further, assume a codec $\mathrm{c} = (\mathrm{enc}, \mathrm{dec})$, where $\mathrm{dec}(\mathrm{enc}(X)) = \tilde{H}$ is an approximation of $H$. The idea is that if $\tilde{H}$ is a sufficient approximation of $H$, then the compressed representation $\mathrm{enc}(X)$ is informative enough of the entire set of tasks $\mathcal{Y}$.  While SVBI still uses task performance as a fidelity measure, the compression model is end-to-end optimized using Head Distillation (HD)~\cite{cde2021, matsubarahnd2019} as the distortion term in the loss function. \Cref{fig:hd} visually explains the loss function. 
We refer to our earlier work~\cite{frankensplit, fool} for a detailed explanation. For this work (i.e., determining adversarial robustness respective bottleneck location depth), it is only relevant that the encoder is shallow with roughly $100,000$-$150,000$ parameters and that the method can significantly reduce bitrate while ensuring task integrity without relying on a labeled dataset. A general downside of lightweight encoders and transmitting information intended to generalize to a broader range of tasks is an increased bitrate relative to deep IB methods.
\section{Problem Statement \& Methodology}
\subsection{Information Bottleneck for Adversarial Robustness} \label{subsubsec:ibfar}
Based on the following two observations, 
we argue that SVBI should still provide a certain level of adversarial robustness but significantly less than DVIB.

First, the depth, i.e., \textit{the large number of stacked layers} up until the bottleneck, may be an essential reason for the efficacy of adversarial robustness using IB-based objectives. Consider an $n$-layered feed-forward neural network as a Markov chain of successive representations $R_{i}, R_{i+1}$~\cite{tishby2015deep}:
\begin{equation}
\label{eq:dnnmarkovchain} 
I(X ; Y) \geq I\left(R_1 ; Y\right) \geq \ldots \geq I\left(R_n ; Y\right) \geq I(\tilde{Y} ; Y)
\end{equation}

\noindent That is, discerning salient from redundant information is part of transforming an input for prediction. A longer sequence of operations permits the network to process the input with more diverse views. 
Therefore, we reason that deeper models may benefit more from the IB objective, as they can learn more varied representations for filtering redundant information (i.e., adversarial noise).

\begin{figure}[htb]
    \centering
    \includegraphics[width=\columnwidth]{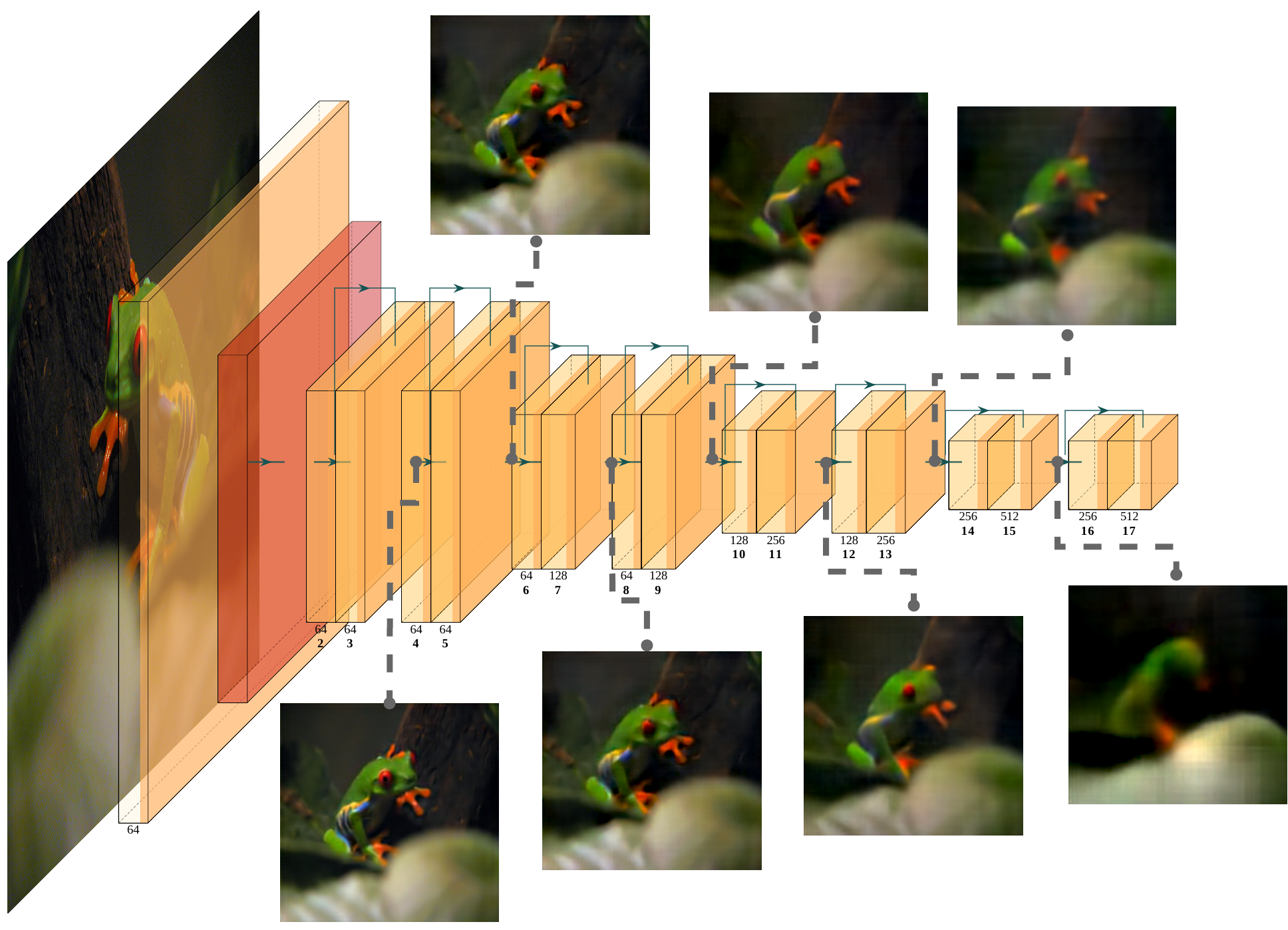}
    \caption{Filtering Redundant Information for the ImageNet classification task as the network transforms features. For each layer, we trained a separate reconstruction network.}
    \label{fig:waddap}
\end{figure}

Second is the \textit{task specifity} of the objective. Consider a visual illustration of the 
information path equation in  \Cref{fig:waddap}.
The frog subset of ImageNet distinguishes between \textit{Tree Frogs}, \textit{Bullfrogs}, and \textit{Tailed Frogs}. Since these frog species have distinct figures and dominant colors, the more delicate characteristics of a tree frog are redundant for ImageNet classification. In SVBI, we place the bottleneck in the first or second marker region, whereas in DVIB, we place it around the last marker. Clearly, when the target task is specifically ImageNet, there is still a considerable amount of redundancy. In other words, when we use an IB-based objective that aims to generalize to a wider range of tasks, there is more ambiguity to exploit.
\subsection{Attack Surface of Task-Oriented Communication Systems} \label{subsubsec:incattacksurface}
\begin{figure}[ht]
    \centering
    \includegraphics[width=\columnwidth]{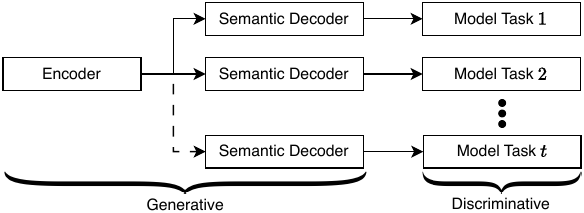}
    \caption{A simplified overview on a Task-Oriented Communication System. Currently envisioned systems rely on generative models to encode, transmit and decode salient information for downstream tasks using discriminative models.}
    \label{fig:sem_com_system}
\end{figure}

\Cref{fig:sem_com_system} illustrates a simplified task-oriented communication system that relies on some form of generative method for compression that can extract and recover salient information. Before passing the input to a discriminative prediction model, we process it with a generative compression model. We argue that even if training the goal-oriented neural codec with an IB-based objective improves adversarial robustness against attacks intended for discriminative tasks, we are still increasing the attack surface of our overal communication system due to the generative components. Therefore, even if the system uses the generative component only for extracting and recovering salient information, exploiting generative components should still be possible, such that it compromises the entire system.
\subsection{Adversarial Attacks and Image Perturbations} \label{subsec:imgperturb}
\begin{figure}[ht]
    \centering
    \includegraphics[width=\columnwidth]{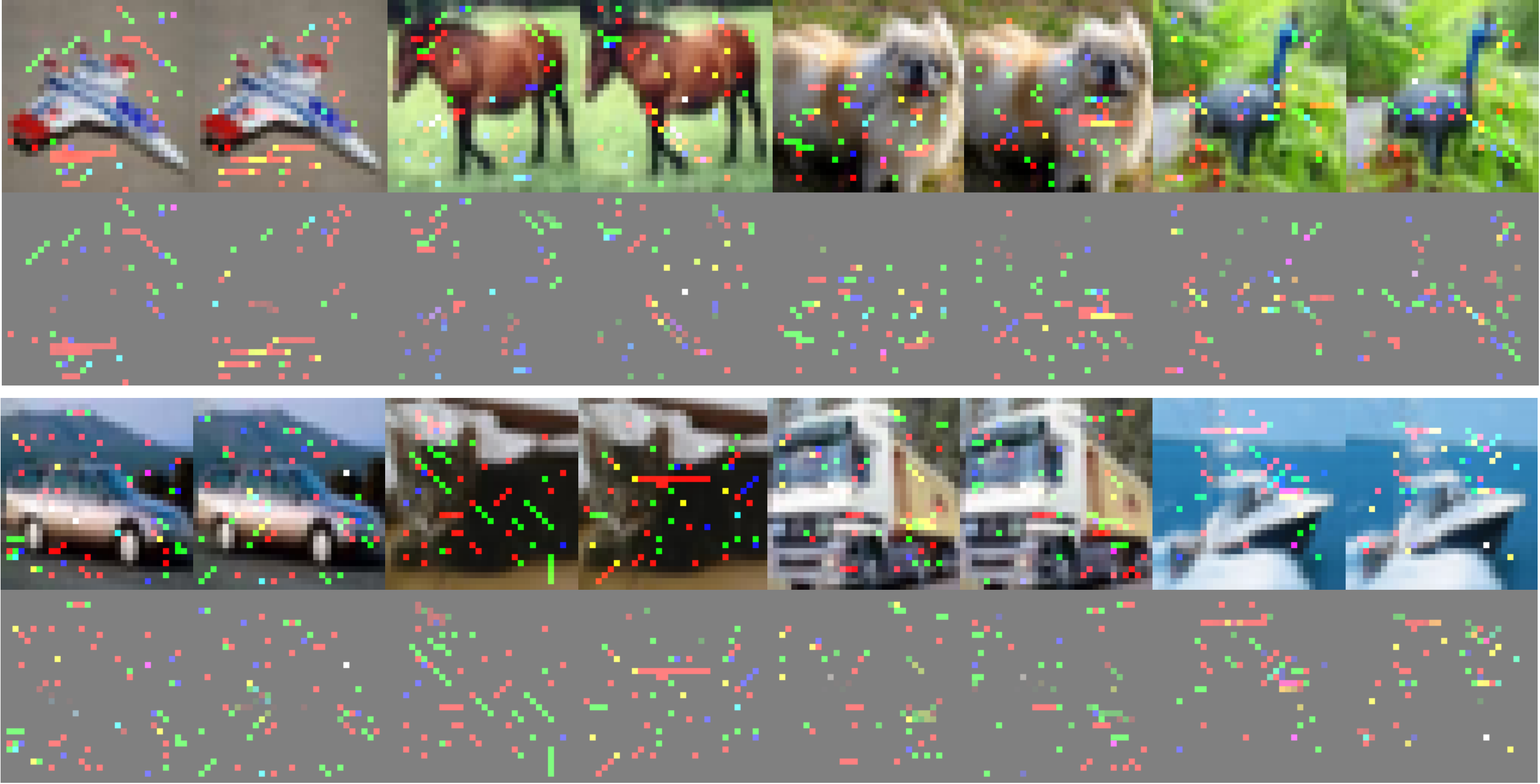}
    \caption{In pairs, comparing JSMA (left) with JSMAOnePixel variant (right).}
    \label{fig:JSMAvsOnePixel}
\end{figure}

We generate adversarial samples using the torchattacks \cite{kim2020torchattacks} library. Except for the Tabacof attack, we create samples for \textit{CIFAR-10},  \textit{SVHN}, and \textit{ImageNet64} (i.e., downsampled ImageNet but still using all original $1000$ classes). Notably, we choose JSMA as it may provide a different perspective on model vulnerability by perturbing specific input features. However, JSMA has high memory requirements, which we cannot accommodate with our limited resources for ImageNet64. Therefore, we implement a modified version of JSMA (JSMAOnePixel) that is inspired by \cite{ADVJS}. The OnePixel variant identifies only a single pixel with the highest impact on each iteration. Still, as \Cref{fig:JSMAvsOnePixel} exemplifies, the final perturbed image is comparable between JSMA and JSMAOnePixel. 
We have verified that applying JSMAOnePixel yields comparable results as reported in \Cref{sec:eval} with lower dimensional samples.

\section{Evaluation} \label{sec:eval}
\subsection{Adversarial Attacks and Image Perturbations} \label{subsec:advattacks}
The aim is to design experiments that yield adequate empirical evidence to conclude the baseline robustness we may expect for the types of compression models used for task-oriented communication. We perform the attacks described in  \Cref{subsec:attackdesc} for each model and task separately using the datasets described in \Cref{subsec:imgperturb}. This is with the exception of  the Tabacof attack, where we use MNIST for simplicty as the purpose is to demonstrate the widened attack surface incurred by task-oriented communication. 
\subsection{Training Models with (Shallow) Bottlenecks}
We train three sets of models, i.e., baseline models with standard log-loss, models with a shallow bottleneck (SVBI), and models with a deep bottleneck (DVIB). We perform DVIB and SVBI as described in \cite{singh2020} and \cite{frankensplit}, respectively. For DVIB, we place a bottleneck at the penultimate layer and use a log-loss for the distortion term in the objective function. For SVBI, we follow the ``blueprint'' encoder design that replaces the layers until the first high-level block of the network (roughly $1\%$ of the total model parameters) with a small variational autoencoder. We experimentally determine the lowest possible bitrate for both bottleneck approaches without sacrificing prediction performance.

\Cref{tab:training_perf} summarizes the model performances.
\begin{table}[htb]
\caption{Models Prediction and Compression Performance}
\label{tab:training_perf}
\resizebox{\columnwidth}{!}{%
\begin{tabular}{lrrr}
\multicolumn{1}{r}{Dataset} & Acc@1 {[}\%{]}    & Bpp (SVBI) & Bpp (DVIB) \\ \hline
MNIST                       & 97.36 $\pm$  1.77 & 0.0829     & 0.0161     \\
\rowcolor[HTML]{EFEFEF} 
CIFAR-10                    & 85.25 $\pm$ 1.40  & 0.5677     & 0.0308     \\
SVHN                        & 94.04 $\pm$ 0.69  & 0.4321     & 0.0086     \\
\rowcolor[HTML]{EFEFEF} 
ImageNet64                  & 49.36 $\pm$ 1.12  & 1.2673     & 0.0115     \\
\hline
\end{tabular}%
}
\end{table}
The bits per pixel (bpp) is a lower bound we have empirically determined for a bottleneck injected model to perform (near-)lossless prediction as defined in \cite{frankensplit, fool}.  Naturally, DVIB has much lower bitrates for reasons described in \Cref{subsubsec:ibfar}.
\subsection{Comparing Bottleneck Placements}
\Cref{tab:resilience_all} summarizes the effect on the adversarial samples represented as percentage points (lower is better). Unsurprisingly, base models trained using a standard log-loss have a significant drop in accuracy. Relative to the accuracy on the unperturbed dataset (\Cref{tab:training_perf}), all attacks completely tank the model performance. 
In particular, for the SVHN task the performance is at times worse than random guessing. 
As conjectured in \Cref{subsubsec:ibfar}, SVBI generally provides less adversarial robustness than DVIB across all datasets. Notably, task complexity apparently influences the gap in adversarial robustness between SVBI and DVIB. Still, SVBI exhibits considerably higher adversarial robustness over the baseline. 
Additionally, notice that the model depth on the base model does not considerably affect adversarial robustness.  However, for the DVIB model, depth seems to correlate positively with adversarial robustness.
Presumably, since deeper models have longer information paths,  end-to-end training models with an IB objective have more opportunities to discard information that does not contribute to task performance gradually. 
\begin{table*}[htb]
\caption{Comparing Prediction Performance Decrease (\% points; lower is better) between Objectives.}
\label{tab:resilience_all}
\resizebox{\textwidth}{!}{%
\begin{tabular}{l|lrrr|rrr|rrr}
                                                  &                              & \multicolumn{3}{c|}{CIFAR-10}                                                                       & \multicolumn{3}{c|}{SVHN}                                                                           & \multicolumn{3}{c}{ImageNet64}                                                                      \\ \hline
Model                                             & Attack                       & Base                            & SVBI                            & DVIB                            & Base                            & SVBI                            & DVIB                            & Base                            & SVBI                            & DVIB                            \\ \hline
                                                  & FGSM                         & 74.5621                         & 48.7298                         & 39.513                          & 69.9521                         & 55.8298                         & 48.5728                         & 37.7602                         & 31.8935                         & 28.9807                         \\
                                                  & \cellcolor[HTML]{EFEFEF}EAD  & \cellcolor[HTML]{EFEFEF}85.5256 & \cellcolor[HTML]{EFEFEF}9.6447  & \cellcolor[HTML]{EFEFEF}8.8592  & \cellcolor[HTML]{EFEFEF}89.9427 & \cellcolor[HTML]{EFEFEF}19.8432 & \cellcolor[HTML]{EFEFEF}13.5824 & \cellcolor[HTML]{EFEFEF}35.4344 & \cellcolor[HTML]{EFEFEF}9.2381  & \cellcolor[HTML]{EFEFEF}8.0993  \\
                                                  & C\&W                         & 87.0232                         & 7.7732                          & 6.7682                          & 92.2149                         & 22.9992                         & 18.3259                         & 37.9903                         & 12.5742                         & 10.2117                         \\
\multirow{-4}{*}{ResNet-18}                       & \cellcolor[HTML]{EFEFEF}JSMA & \cellcolor[HTML]{EFEFEF}87.6210 & \cellcolor[HTML]{EFEFEF}20.7807 & \cellcolor[HTML]{EFEFEF}17.5784 & \cellcolor[HTML]{EFEFEF}91.5810 & \cellcolor[HTML]{EFEFEF}14.2348 & \cellcolor[HTML]{EFEFEF}11.1283 & \cellcolor[HTML]{EFEFEF}36.3821 & \cellcolor[HTML]{EFEFEF}11.1868 & \cellcolor[HTML]{EFEFEF}10.1935 \\ \hline
                                                  & FGSM                         & 68.5621                         & 42.8942                         & 34.0803                         & 68.2679                         & 53.2118                         & 45.1977                         & 39.6985                         & 28.9273                         & 23.1021                         \\
                                                  & \cellcolor[HTML]{EFEFEF}EAD  & \cellcolor[HTML]{EFEFEF}85.1400 & \cellcolor[HTML]{EFEFEF}9.1258  & \cellcolor[HTML]{EFEFEF}7.8592  & \cellcolor[HTML]{EFEFEF}89.3852 & \cellcolor[HTML]{EFEFEF}18.0232 & \cellcolor[HTML]{EFEFEF}11.4375 & \cellcolor[HTML]{EFEFEF}32.6361 & \cellcolor[HTML]{EFEFEF}7.0377  & \cellcolor[HTML]{EFEFEF}4.8375  \\
                                                  & C\&W                         & 88.9231                         & 7.2009                          & 6.5408                          & 93.3284                         & 18.0931                         & 15.3259                         & 34.1083                         & 9.9654                          & 5.4281                          \\
\multirow{-4}{*}{ResNet-50}                       & \cellcolor[HTML]{EFEFEF}JSMA & \cellcolor[HTML]{EFEFEF}85.1010 & \cellcolor[HTML]{EFEFEF}19.6333 & \cellcolor[HTML]{EFEFEF}16.0549 & \cellcolor[HTML]{EFEFEF}90.2838 & \cellcolor[HTML]{EFEFEF}13.9125 & \cellcolor[HTML]{EFEFEF}10.1283 & \cellcolor[HTML]{EFEFEF}34.4847 & \cellcolor[HTML]{EFEFEF}10.1351 & \cellcolor[HTML]{EFEFEF}5.5213  \\ \hline
\multicolumn{1}{c|}{}                             & FGSM                         & 69.3189                         & 40.8912                         & 32.1534                         & 66.2082                         & 40.4817                         & 42.9004                         & 38.4451                         & 24.0620                         & 20.9997                         \\
\multicolumn{1}{c|}{}                             & \cellcolor[HTML]{EFEFEF}EAD  & \cellcolor[HTML]{EFEFEF}87.6557 & \cellcolor[HTML]{EFEFEF}8.0322  & \cellcolor[HTML]{EFEFEF}7.8592  & \cellcolor[HTML]{EFEFEF}88.3294 & \cellcolor[HTML]{EFEFEF}16.4385 & \cellcolor[HTML]{EFEFEF}9.5729  & \cellcolor[HTML]{EFEFEF}32.4148 & \cellcolor[HTML]{EFEFEF}6.1124  & \cellcolor[HTML]{EFEFEF}3.0489  \\
\multicolumn{1}{c|}{}                             & C\&W                         & 87.4633                         & 7.1819                          & 6.0018                          & 92.1923                         & 17.3284                         & 12.2482                         & 38.0200                         & 8.7985                          & 2.9663                          \\
\multicolumn{1}{c|}{\multirow{-4}{*}{ResNet-101}} & \cellcolor[HTML]{EFEFEF}JSMA & \cellcolor[HTML]{EFEFEF}86.8781 & \cellcolor[HTML]{EFEFEF}19.439  & \cellcolor[HTML]{EFEFEF}15.2608 & \cellcolor[HTML]{EFEFEF}94.2933 & \cellcolor[HTML]{EFEFEF}11.2814 & \cellcolor[HTML]{EFEFEF}7.9833  & \cellcolor[HTML]{EFEFEF}36.6825 & \cellcolor[HTML]{EFEFEF}10.0382 & \cellcolor[HTML]{EFEFEF}1.4762  \\ \hline
\end{tabular}%
}
\end{table*}

\subsection{Analyzing Pixel Perturbations}
We observe that IB-based objectives exhibit stronger robustness against attacks that focus on a small subset of salient pixels with strong intensity than attacks that perturb many pixels with smaller intensity. Moreover, similarly to the original work on deep variational IB~\cite{deepinformationbottleneck}, we observe that attacks targeting IB-based models perturb pixels considerably more than non-IB-based models. Nevertheless, since relative values align across all models (i.e., attacks behave comparably regardless of the model depth or objective), the following reports average values due to space constraints. 

\begin{figure}[ht]
    \centering
    \includegraphics[width=\columnwidth]{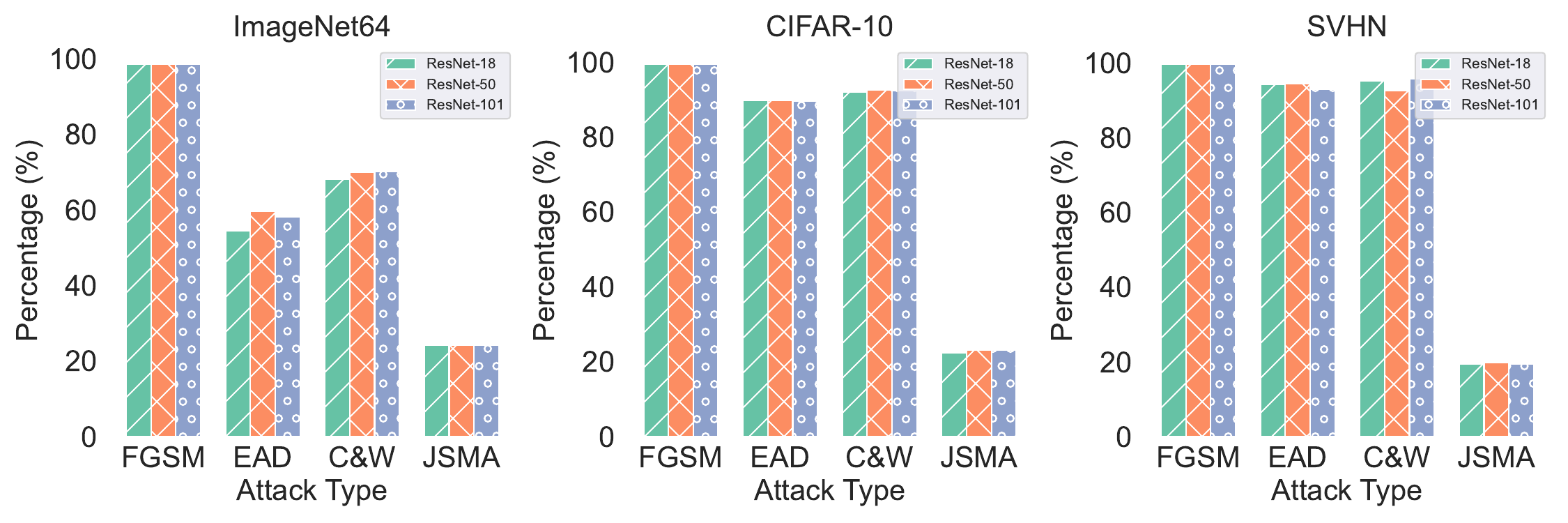}
    \caption{Average Percentage of pixels perturbed by an adversarial attack. More complex tasks tend to have more salient pixels.}
    \label{plot:l0dist}
\end{figure}

\begin{figure}[ht]
    \centering
    \includegraphics[width=\columnwidth]{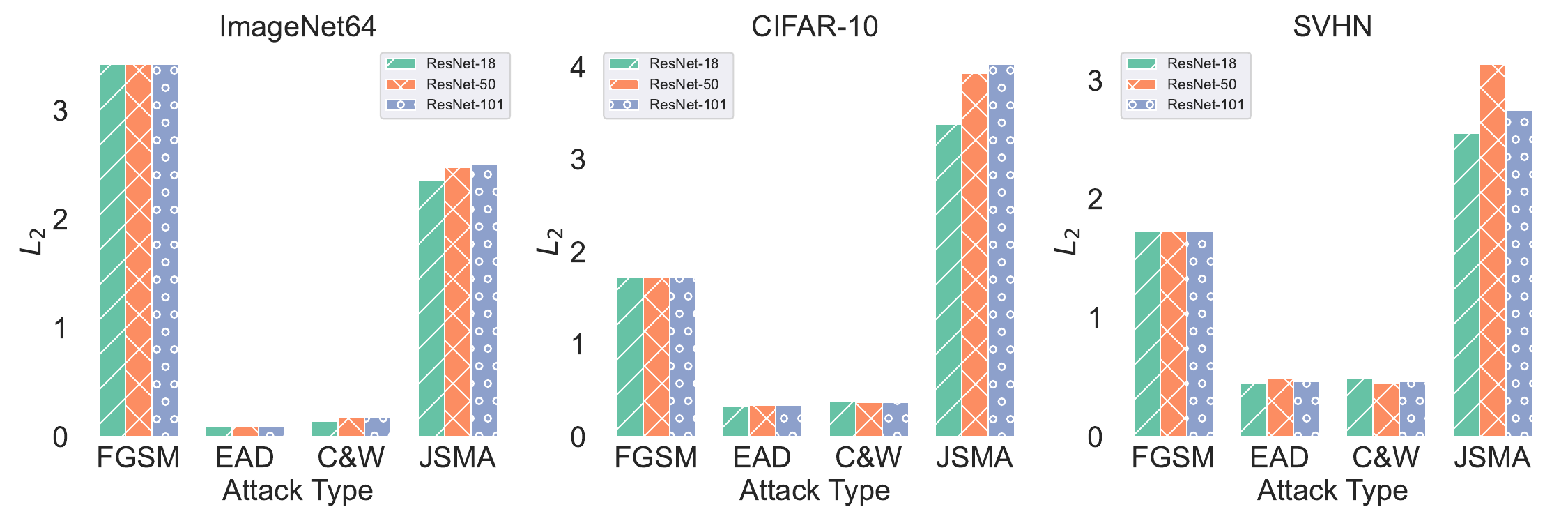}
    \caption{The average $L_{2}$ measures the magnitude of perturbations. FSGSM and JSMA incur considerably higher perturbation than EAD and C\&W.}
    \label{plot:l2dist}
\end{figure}

\begin{figure}[ht]
    \centering
    \includegraphics[width=\columnwidth]{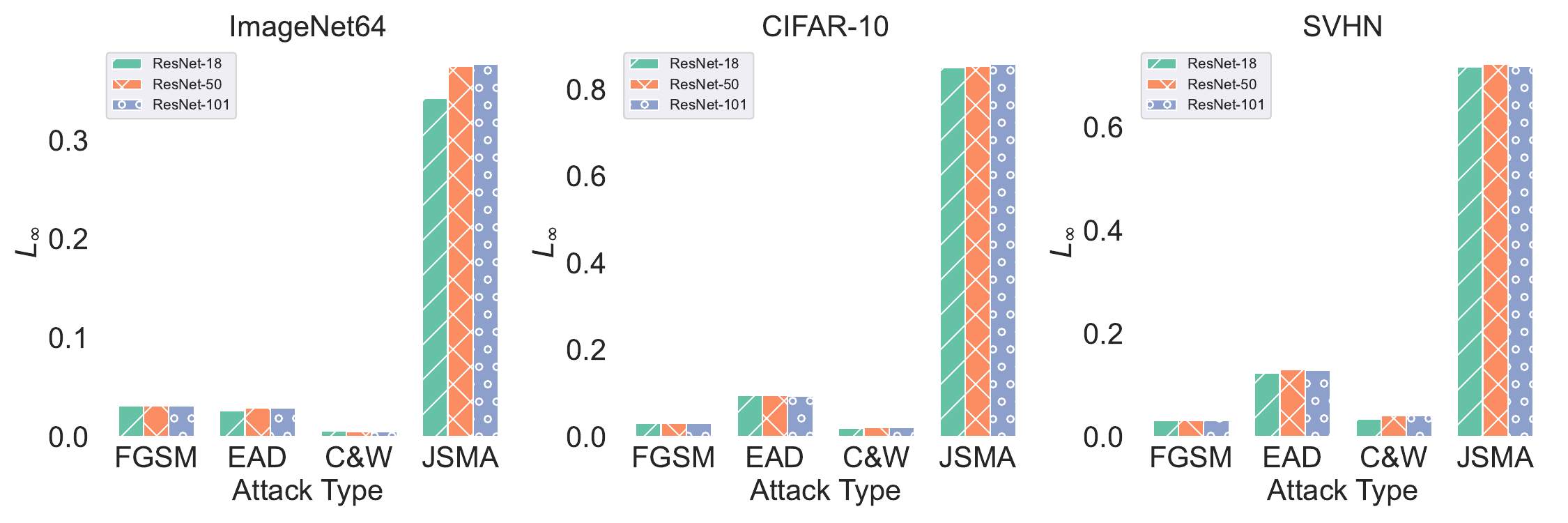}
    \caption{Average $L_{\infty}$ measure to quantify the magnitude of perturbation. JSMA perturbs a small number of pixels with high intensity.}
    \label{plot:linfdist}
\end{figure}

\begin{figure}[ht]
    \centering
    \includegraphics[width=\columnwidth]{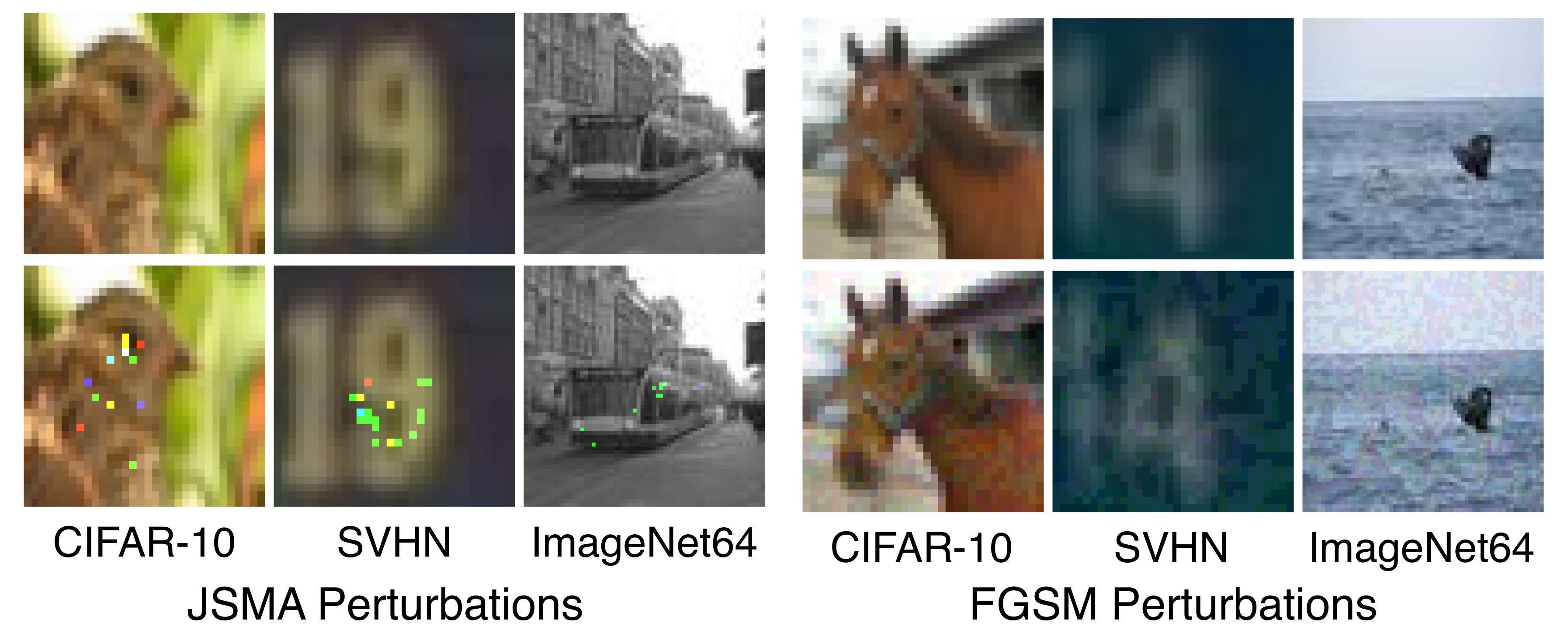}
    \caption{Comparing magnitudes of pixels between JSMA and FGSM.}
    \label{fig:jsma_vs_fgsm}
\end{figure}

\Cref{plot:l0dist} visualizes the $L_{0}$ norm by attack averaged over test sets, i.e., it measures how many pixels an attack has perturbed. 
While FGSM perturbs nearly all pixels, JSMA only perturbs roughly $20\%$ of the pixels. More interestingly, EAD and C\&W perturb fewer pixels for ImageNet than for the simpler tasks. Generally, more complex tasks with many labels rely on more fine-grained information, where just a small subset of salient pixels can influence the decision boundaries. \Cref{plot:l2dist} summarizes the magnitude of perturbations. 
Notice that JSMA has a larger total magnitude in total perturbation than FGSM despite JSMA focusing on a smaller subset of pixels. The reason becomes apparent when examining the $L_\infty$ norm in \Cref{plot:linfdist}.
JSMA is more ``pixel-efficient'' by focusing on the most salient pixels but relies on high-magnitude perturbations. \Cref{fig:jsma_vs_fgsm} visually compares JSMA and FGSM. While FGSM perturbs a large number of pixels, they are only faintly visible. Conversely, JSMA has clearly visible perturbations.
This observation is consistent with the general objective of goal-oriented communication, which is to focus on the most salient information. Therefore, it may be reasonable to emphasize evaluating defense strategies for task-oriented communication against less perceptible attacks. 
\subsection{Targeted Autoencoder Attack}
\begin{figure}[ht]
    \centering
    \includegraphics[width=\columnwidth]{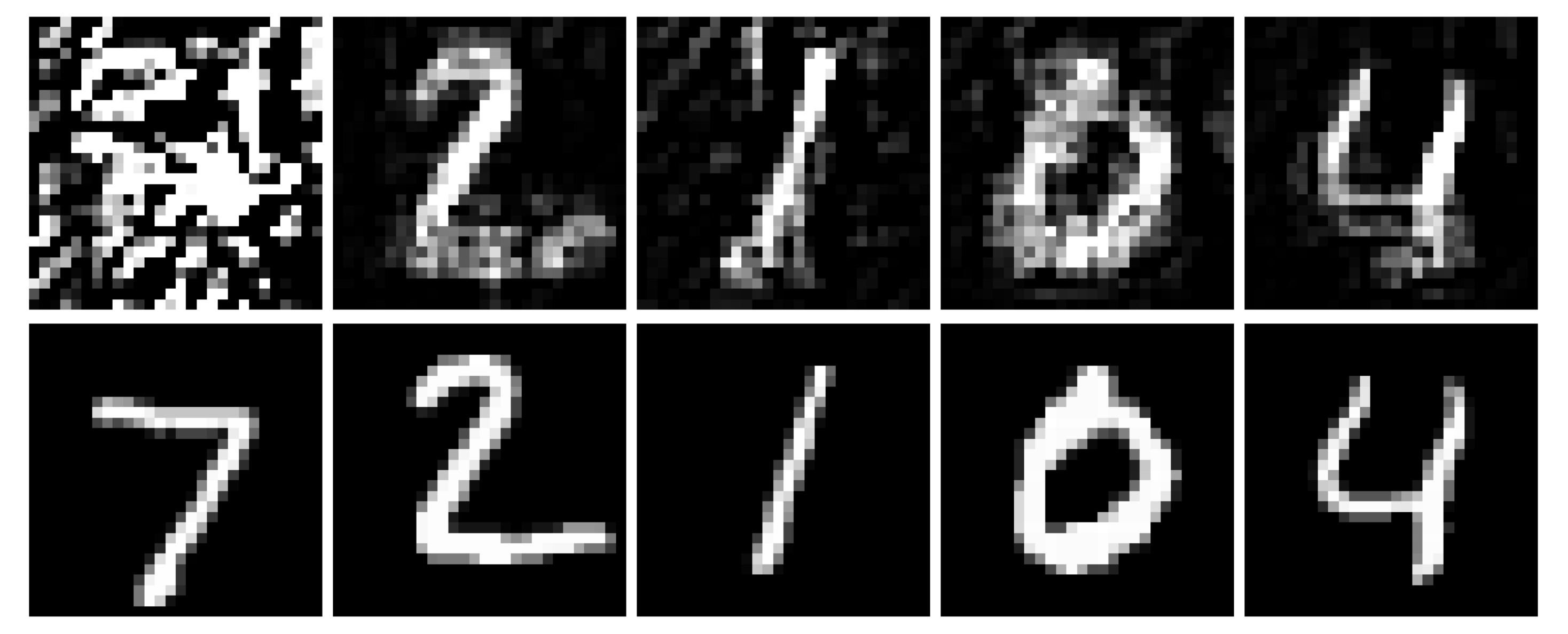}
    \caption{Base images (bottom) and corresponding perturbation using the Tabacof attack prepared against ResNet-18 with DVIB (top). The target label is ``1''. Examples depicting the numbers two and four contain the most clearly visible perturbations to match this target. The leftmost example showcases a ``failed'' attack, where the network will likely missclassify the input, but not hit the intended target.}
    \label{fig:tabacof_viz}
\end{figure}

As described in \Cref{subsubsec:incattacksurface} task-oriented communication networks are powered by generative models for communication and discriminative models for high-level downstream tasks, which incrases the attack surface. We show this by performing the Tabacof~\cite{Tabacof2016AdversarialIF} attack described in \Cref{subsubsec:tabacof} and summarize the results in \Cref{tab:tabacof-2}. 
\begin{table}[htb]
\caption{Tabacof attack}
\label{tab:tabacof-2}
\resizebox{\columnwidth}{!}{%
\begin{tabular}{l|ll|ll}
\multicolumn{1}{c}{}                             & \multicolumn{2}{c}{Base}    & \multicolumn{2}{c}{DVIB} \\\hline 
Model                        & Acc@1  & \#\,Hits                          & Acc@1  & \#\,Hits\\\hline
\cellcolor[HTML]{EFEFEF}Resnet-18                    & \cellcolor[HTML]{EFEFEF}61.6  & \cellcolor[HTML]{EFEFEF}927                         &  \cellcolor[HTML]{EFEFEF}52.26 & \cellcolor[HTML]{EFEFEF}1802  \\
Resnet-50                    & 76.57  & 879                        & 72.66  & 1126  \\
\cellcolor[HTML]{EFEFEF}Resnet-101                   & \cellcolor[HTML]{EFEFEF}33.17  & \cellcolor[HTML]{EFEFEF}448                        &\cellcolor[HTML]{EFEFEF}34.73   & \cellcolor[HTML]{EFEFEF}1641  \\\hline
\end{tabular}%
}
\end{table}
We choose the label ``1'' as the target, and the \textit{hits} column indicates how often the model has predicted ``1'' after the attack. Since the models have near-perfect accuracy on MNIST, and the test set has $10\,000$ samples that are uniformly distributed, we can infer the efficacy of the attack by the increase of predictions of ``1''.  \Cref{fig:tabacof_viz} shows an example with a curated sample of perturbed images.
ResNet-101 is noticeably less robust toward the attack. We explain the discrepancy by the simplicity of the task and dataset size. Since ResNet-101 is significantly larger, the model may have been fitted to the samples, making it more susceptible to even slight perturbations.
Still, when comparing the performance of ResNet-18 and ResNet-50 shows that the DVIB model is considerably \textit{less robust} than the baseline model. Notably, all DVIB models have a substantial increase in predicting the attack target label, indicative of the attack's efficacy.
\section{Conclusion} \label{sec:conclusion}
In this work we have investigated the role of IB-based objectives for task-oriented communication systems and its implication for adversarial robustness. We have shown that while such approaches provide a degree of resilience against attacks targeting downstream tasks, the reliance on generative models for extracting and recovering salient bits introduces a new attack surface. Naturally, covering the entire landscape of adversarial attacks and adjusting them for all types of semantic communication systems is not within the scope of this work. 
Lastly, we find an interesting direction for future work are methods that quantify the trade-off of how well the salient information generalizes and its adversarial robustness as an objective function for end-to-end optimization of goal-oriented neural codecs.

\section*{Acknowledgment}
We thank Alexander Knoll for providing us with the hardware infrastructure and Valentin Flunkert for his support.
\bibliographystyle{IEEEtran}
\bibliography{main}
\end{document}